\documentclass[sigconf,natbib=true]{acmart}
\AtBeginDocument{%
  }

\setcopyright{acmlicensed}
\copyrightyear{2025}
\acmYear{2025}
\acmDOI{XXXXXXX.XXXXXXX}
\acmConference[SSTD'25]{19th International Symposium on Spatial and Temporal Data}{25-27 August, 2025}{Osaka, Japan}

\acmISBN{978-1-4503-XXXX-X/2018/06}

\usepackage{xcolor}
\usepackage{hyperref}
\usepackage{caption}
\usepackage{subcaption}
\usepackage{balance}
\usepackage{amsmath} 
\usepackage{tabularx}
\usepackage{multirow}
\usepackage{booktabs}
\usepackage{siunitx} 
\usepackage{makecell}

\setcitestyle{numbers,sort&compress}

\begin{document}

\title[Physics-Informed Vessel Trajectory Prediction: A Finite Difference Approach]%
{Physics-Informed Neural Networks for Vessel Trajectory Prediction: %
Learning Time-Discretized Kinematic Dynamics~via~Finite~Differences}

\author{Md Mahbub Alam}
\email{mahbub.alam@dal.ca}
\orcid{0000-0003-0756-264X}
\affiliation{%
  \institution{Dalhousie University}
  \city{Halifax}
  \state{NS}
  \country{Canada}
}

\author{Amilcar Soares}
\email{amilcar.soares@lnu.se}
\orcid{0000-0001-5957-3805}
\affiliation{%
  \institution{Linnaeus University}
  \city{Växjö}
  \state{}
  \country{Sweden}
}

\author{José F. Rodrigues-Jr}
\email{junio@icmc.usp.br}
\orcid{0000-0001-8318-1780}
\affiliation{%
  \institution{University of Sao Paulo}
  \city{Sao Carlos}
  \state{SP}
  \country{Brazil}
}

\author{Gabriel Spadon}
\authornote{Corresponding Author}
\email{spadon@dal.ca}
\orcid{0000-0001-8437-4349}
\affiliation{%
  \institution{Dalhousie University}
  \city{Halifax}
  \state{NS}
  \country{Canada}
}

\renewcommand{\shortauthors}{Alam et al.}

\begin{abstract}
Accurate vessel trajectory prediction is crucial for navigational safety, route optimization, traffic management, search and rescue operations, and autonomous navigation. Traditional data-driven models lack real-world physical constraints, leading to forecasts that disobey vessel motion dynamics, such as in scenarios with limited or noisy data where sudden course changes or speed variations occur due to external factors. To address this limitation, we propose a Physics-Informed Neural Network (PINN) approach for trajectory prediction that integrates a streamlined kinematic model for vessel motion into the neural network training process via a first- and second-order, finite difference physics-based loss function. This loss function, discretized using the first-order forward Euler method, Heun's second-order approximation, and refined with a midpoint approximation based on Taylor series expansion, enforces fidelity to fundamental physical principles by penalizing deviations from expected kinematic behavior. 
We evaluated PINN using real-world AIS datasets that cover diverse maritime conditions and compared it with state-of-the-art models. Our results demonstrate that the proposed method reduces average displacement errors by up to $32\%$ across models and datasets while maintaining physical consistency. 
These results enhance model reliability and adherence to mission-critical maritime activities, where precision translates into better situational awareness in the oceans. 
\end{abstract}




\maketitle

\section{INTRODUCTION}

Enhancing situational awareness on ocean transportation is key for preventing collisions, a risk that has grown significantly due to the substantial increase in maritime traffic in recent years~\cite{Wan2018, Anto2023, alam2024enhancing}. Beyond improving situational awareness, the accurate prediction of future vessel movements is critical for a wide range of maritime applications. These include enabling efficient navigation through route optimization, reducing fuel consumption and transit times~\cite{yang2024graph}; supporting proactive traffic management to alleviate congestion and improve safety in busy waterways~\cite{zissis2016real, xiao2019traffic}; informing timely and effective responses during maritime emergencies, such as search and rescue operations; aiding in the detection and tracking of vessels involved in illicit activities, thereby contributing to maritime security~\cite{ray2020mobility}; and providing vital input for the safe and reliable operation of autonomous vessels~\cite{Cuomo2022}. Trajectory prediction has emerged as an extensively explored research area in the maritime domain, driving innovation across applications~\cite{Zhang2022, Li2023}.

Innovative approaches are being developed to address the challenges of accurate trajectory prediction, particularly using machine learning techniques. One significant recent advancement in machine learning is the Physics-Informed Neural Networks (PINNs), well-suited for applications governed by physical principles~\cite{Cuomo2022, Toscano2024}, such as vessel trajectory prediction. Unlike purely data-driven models that rely solely on observed data, PINNs embed domain-specific physical laws into the neural network training process, improving data efficiency, enhancing generalization, and ensuring physically consistent predictions. As vessel movement follows kinematic principles, PINNs can integrate these dynamics into the model, leading to accurate and reliable forecasts.

Over the past decade, maritime trajectory prediction research has shifted from relying solely on physical principles to adopting data-driven models, leveraging classical machine learning and, more recently, deep learning~\cite{xiao2019traffic, Zhang2022, Li2023}. This transition has been particularly impactful in recent years, with deep learning models achieving significant improvements in prediction accuracy. These advancements began with RNN-based architectures such as LSTMs and GRUs~\cite{chen2023tdv, Spadon2024, slaughter2024vessel}, progressed to CNNs~\cite{shin2024deep}, and now encompass more advanced models such as Temporal Convolutional Networks (TCNs)~\cite{lin2023ship}, Graph Neural Networks (GNNs)~\cite{yang2024graph}, and Transformers~\cite{nguyen2024transformer, lin2025hdformer}. The initial objective was to improve prediction accuracy by using large datasets and introducing novel approaches through architectural changes or model fusion~\cite{jiang2024stmgf, wu2024vessel}. Additionally, some approaches incorporate a distance-based loss function to minimize the displacement between predicted and actual trajectories~\cite{guo2025vessel, zhang2025goal}. However, in scenarios involving limited, sparse, or noisy data, the accuracy of these models remains limited, especially when sudden course changes or speed variations arise due to environmental factors. PINNs offer a promising solution by integrating physical laws with observed data, enabling them to achieve more accurate and reliable predictions, even under challenging conditions.

Moreover, since the International Maritime Organization (IMO) mandated the adoption of the Automatic Identification System (AIS) in 2004~\cite{Edition2004_consolidated} and its subsequent satellite integration in 2008~\cite{ORBCOMM_2016_Satellite-AIS}, AIS data has become integral to maritime research. Particularly, dynamic AIS messages containing crucial kinematic information such as latitude, longitude, speed over ground (SOG), and course over ground (COG) have been extensively leveraged for diverse applications, including vessel trajectory prediction. While predominantly data-driven approaches have utilized these core AIS features as inputs for prediction models~\cite{Zhang2022}, some methods have further incorporated derived kinematic attributes — including acceleration, jerk, COG rate, and bearing — to capture complex vessel mobility patterns~\cite{alam2024enhancing}. However, a critical gap remains: the lack of integrated modeling of the kinematic constraints inherently embedded in AIS data, which fundamentally governs real-world vessel dynamics. Our work addresses this gap by integrating vessel kinematic constraints into data-driven models through a PINN framework.

This paper proposes a PINN framework that leverages first-order and second-order finite-difference physics loss functions to enhance vessel trajectory prediction using AIS data.
These loss functions, derived from a simplified kinematic model of vessel motion, enforce physical consistency by penalizing deviations from physically expected kinematic behavior. 
The kinematic model was discretized using a first-order forward Euler method and Heun's second-order approximation, implemented with forward finite differences. 
To further improve accuracy, a midpoint approximation based on a first-order Taylor series expansion was also incorporated into the calculation of expected displacement. Importantly, the proposed PINN framework is model-agnostic, allowing seamless integration with state-of-the-art deep learning architectures, including RNNs, CNNs, ConvLSTM, TCNs, and Transformers, for vessel trajectory prediction. We conducted a comprehensive evaluation by varying model complexity, prediction horizons, and the order of numerical approximations. Moreover, we factored in the complexity of navigational areas when assessing prediction accuracy.

The remainder of this paper is organized as follows to present our contributions. Section 2 reviews recent related work; Section 3 outlines the methodology and proposal formulation; Section 4 presents and analyzes the experimental results derived from the comparison with baseline models; and Section 5 concludes with remarks and potential future research directions.

\section{RELATED WORK}
With the surge of studies in physics-driven machine learning, several review studies explored its applicability across domains~\cite{Cuomo2022, Toscano2024}. However, in the intersection of ocean mobility, maritime system reliability, and shipping operations, the meeting between PINNs and their applications remains underexplored. This gap exists largely due to the lack of direction and synthesis needed to support future research, define sound baselines, and map the landscape of physics-guided learning in ship movement and operations. To build towards such a foundation, we categorize the literature into groups based on how physical knowledge is integrated into models:
\begin{enumerate}
    \item \textit{\textbf{Physics as Structure:}} models that directly embed physical laws ({\it i.e.}, equations) into their architectures, ensuring that vessel trajectories conform to established physical principles;

    \item \textit{\textbf{Physics as Direction:}} hybrid methods that enrich data-driven models with features, priors, or multi-modal signals, blending domain knowledge with learning from data; and,

    \item \textit{\textbf{Physics as Correction:}} approaches that use physics to regularize or correct predictions, such as guiding generative processes or refining numerical outputs to achieve realistic and reliable predictions.
\end{enumerate}

Within such categories, which are detailed subsequently, it is possible to observe that current models often incorporate only partial physics. Many physics-informed models use environmental data or simple motion constraints but do not yet capture the full complexity of ship dynamics ({\it e.g.}, hull hydrodynamics and six-degree-of-freedom motions\footnote{~\textbf{6-DoF:} \textit{Surge} (forward/backward), \textit{Sway} (left/right), \textit{Heave} (up/down), \textit{Roll} (rotation around X-axis), \textit{Pitch} (rotation around Y-axis), and \textit{Yaw} (rotation around Z-axis).}) in the learning process. Fully integrating high-fidelity naval architecture models into neural networks remains to be seen due to computational complexity, increased computational cost, and the need for expert knowledge. Working with AIS data poses additional constraints because the data inherently lack detailed kinematic and dynamic measurements.

This paper serves as a primer on applying PINNs with AIS data for ocean mobility and its related applications. Current AIS-based research predominantly employs 2-DoF approaches, ignoring the Z-axis, although specific scenarios like shallow waters or vessel draft sometimes warrant 3-DoF models. The full integration of 6-DoF motion remains limited. While research continues to seek better solutions, this work takes a significant step by introducing the first theoretical framework to properly integrate kinematic physics into neural networks for trajectory modeling, aiming to enhance model applicability, usability, and accuracy.\\

\noindent(1)~\textit{Structure -- \textbf{Models Constrained by Equations}}\\

Zhao {\it et al.} (2024)~\cite{Zhao2024} propose a framework for Autonomous Underwater Vehicles (AUV) that integrates PINNs with classical dynamic equations to capture the full 6-DoF motion of AUVs. The study embeds the spatial maneuvering motion equations directly into the loss function of a fully connected neural network, which is trained via a multi-step iterative process using a fourth-order Runge–Kutta scheme for time integration. 
Simulation experiments demonstrate that PINN yields stable, accurate, generalized long-term motion prediction even with limited training data. Field tests on a micro-AUV confirm effective trajectory tracking with control errors below one degree.

Papandreou {\it et al.} (2025)~\cite{Papandreou2025} propose an interpretable model that augments traditional physics-based motion prediction by optimizing key hydrodynamic parameters via a constrained nonlinear least squares method. The model integrates a 3-DoF\footnote{~\textbf{3-DoF:} \textit{Surge} (forward/backward), \textit{Sway} (left/right), and \textit{Heave} (up/down).} physics-based framework (which incorporates rudder and propeller forces and a vessel resistance curve) with data-driven parameter tuning to capture ship-specific behaviors. Eleven parameters, including those governing the resistance polynomial and rudder force coefficients (for both lift and drag), are estimated using synthetic trajectory data derived from realistic ship maneuvers. 
Validation on datasets of two container ships demonstrates that fitted models predict trajectories with $51.6-57.8\%$ higher accuracy and $72.36-89.67\%$ greater consistency compared to conventional baselines.

Mathioudakis {\it et al.} (2025)~\cite{Mathioudakis2025} present a three-dimensional physics‐based model of ship motion prediction tailored for long container vessels. The model integrates dynamic equations for surge, sway, and yaw with hydrodynamic derivative methods to compute forces from control inputs (rudder and propeller) and environmental effects (wind, waves, and currents) via numerical integration. Validation against a baseline maneuvering model and real‑world sea trial data 
demonstrates that the model can replicate vessel trajectories under typical conditions while revealing challenges during transient maneuvers with small rudder angles.\\

\noindent(2)~\textit{Direction -- \textbf{Hybrid and Augmented Learning}}\\


Lang, Wu, and Mao (2024)~\cite{Lang2024} introduce a physics-informed grey-box model for predicting ship speed by combining a physics-based component implemented using PINN and a data-driven component based on XGBoost. The methodology employs a parallel modeling architecture where the PINNs estimate the expected calm water speed from propulsion power and draft using speed-power model tests. In contrast, the XGBoost model predicts the speed reduction under current ocean conditions. Validation using full-scale performance data from a chemical tanker demonstrates that the model achieves approximately 30\% improvement in prediction accuracy over a traditional black-box model and reduces the accumulated error in estimated time of arrival by about $50\%$.

Chen {\it et al.} (2025)~\cite{Chen2025} proposed a hybrid approach that combines an LSTM neural network with a hybrid PSO-GWO ({\it i.e.}, swarm-based meta-heuristic) algorithm for short-time ship trajectory prediction. The model captures vessel motion's nonlinear and time-varying characteristics by leveraging the long-term dependencies in AIS data. The results indicate that the proposal achieves lower mean absolute and squared errors while reducing optimization time and enhancing predictive accuracy and computational efficiency.

Guo et al. (2025)~\cite{Guo2025} introduce the Vessel Influence LSTM model for trajectory prediction, incorporating Vessel Influence Maps to capture the dynamic effects of surrounding vessels. The model integrates vessel motion, environmental, and static factors with the influence of neighboring vessels while also employing Gaussian prediction combined with Monte Carlo dropout to estimate uncertainty. A temporally weighted hybrid loss function is proposed to balance prediction accuracy with uncertainty quantification. 
Experimental results on AIS data from Galveston Bay, US, demonstrate that the model achieves lower mean distance errors than baseline models on standard and unseen test sets, particularly under complex and high-density maritime conditions.

Suo, Ding, and Zhang (2024)~\cite{Suo2024} propose a deep-learning framework based on Mamba for ship trajectory prediction that employs a selective state-space model to process long sequential data efficiently. The model integrates hardware-aware state expansion and a simplified architecture to overcome the limitations of conventional methods such as LSTM, GRU, and Transformer. The model improved prediction accuracy, inference speed, and resource utilization on AIS data from the Beijing–Hangzhou Canal.

Zhao {\it et al.} (2025)~\cite{Zhao2025} propose a deep learning framework that integrates inter-ship interactions and navigational uncertainties into the prediction model. It is built upon an encoder-decoder LSTM architecture that incorporates three key attention-based modules: a Position Attention Block that captures mutual positional influences among vessels, an Information Fusion Block that integrates differential navigation state information to represent uncertainty, and a Global Attention Block that dynamically aligns encoder outputs with the decoder's context. The experimental results on AIS datasets from multiple maritime regions demonstrate that the model outperforms baseline models.

Song et al. (2024)~\cite{Song2024} propose the Transformer Gravity model, a gravity‑inspired deep learning framework to forecast global maritime traffic flows and enhance risk assessments for non-indigenous species spread via ballast water. The model extends traditional gravity formulations by integrating features, such as shipping flux density, geodesic distances, bilateral trade volumes, and graph-based centrality metrics, into a Transformer architecture that effectively captures short‑ and long‑term dependencies in vessel movement. Evaluated on global shipping networks derived from AIS data (2017--2019), this approach demonstrates over $10$\% higher prediction accuracy compared to conventional deep‑gravity and regression models. 
\\

\noindent(3)~\textit{Correction -- \textbf{Generative and Precision-Critical Models}}\\

Zhang {\it et al.} (2025)~\cite{Zhang2025} propose a diffusion probabilistic framework for long-term vessel trajectory imputation that tackles extensive AIS data gaps. Their model uses a pre-trained trajectory embedding block to extract movement patterns and a transformer encoder to condition the reverse denoising process, generating continuous, multi-point imputed trajectories. A physics-guided discriminator enforces kinematic constraints between positional and angular data and curbs cumulative error. Experiments on a real-world AIS dataset demonstrate that this approach recovers long-term vessel trajectories with improved accuracy compared to existing imputation methods.


Ferreira and Campbell (2025)~\cite{Ferreira2025} introduce an RNN architecture for ship trajectory prediction that incorporates a Decimal Preservation (DP) layer to capture minute latitude and longitude variations and mitigate floating-point rounding errors. Against the Ornstein–Uhlenbeck baseline, the proposed architecture reduces the prediction errors of cargo vessels by up to $50\%$ and demonstrates that the DP layer enhances the performance of Elman's RNN, LSTM, and GRU models.

\section{METHODOLOGY}
This section defines the trajectory prediction problem and details the Physics-Informed Neural Network (PINN) architecture.

\subsection{Preliminaries}

\subsubsection*{\textbf{Trajectory}}
A trajectory \( \mathcal{T} \) is a sequence of points generated by a moving object. Each point \( p_t \) in the trajectory represents the object's coordinates \( (x_t, y_t) \) and its kinematic state at time \( t \):
\begin{equation}
    \mathcal{T} = \{p_1, p_2, \dots, p_N \}
    \label{eq:trajectory_definition}
\end{equation}
\noindent where \( N \) is the number of temporal observations in the trajectory. For vessel trajectories, each point \( p_t = (x_t, y_t, v_t, \psi_t, a_t, \dot{\psi}_t) \) is characterized by the following features at time \( t \):
\begin{itemize}
    \item \( x_t \): Latitude (in degrees)
    \item \( y_t \): Longitude (in degrees)
    \item \( v_t \): Speed Over Ground (SOG, in meters per second)
    \item \( \psi_t \): Course Over Ground (COG, in degrees)
    \item \( a_t \): Acceleration (in meters per second squared)
    \item \( \dot{\psi}_t \): rate of change of COG (in degrees per second)
\end{itemize}
\noindent Accordingly, for a point $p_t$, the input feature vector for the neural network model is defined as its transpose $p{_t}^T = \mathbf{x}_t$:
\begin{equation}
    \mathbf{x}_t = [x_t, y_t, v_t, \psi_t, a_t, \dot{\psi}_t]^T
    \label{eq:input_vector}
\end{equation}

\subsubsection*{\textbf{Trajectory Prediction}}
Trajectory prediction is the task of forecasting the future trajectory with respect to a point of interest $p_t$. That is,
\( \mathcal{T}_{\text{pred}} = \{ \hat{p}_{t+1}, \hat{p}_{t+2}, \dots, \hat{p}_{t+H} \} \),
of a moving object, given its observed trajectory
\( \mathcal{T}_{\text{obs}} = \{ p_1, p_2, \dots, p_t \} \).
\( H \) is the prediction horizon, indicating the number of future time steps to be predicted. 

\subsection{PINN Architecture}
\label{ssec:pinn-architecture}

The principle of Physics-Informed Neural Networks (PINNs) is to train a neural network to learn from observed data and satisfy underlying physical equations.
A PINN, denoted \( f_{\theta}(\mathbf{x}_t) \), acts as a function approximator, learns a mapping from the input feature vector $\mathbf{x}_t=p{_t}^T$ to the predicted output
\( \hat{\mathbf{y}}_{t+i} = [\hat{x}_{t+i}, \hat{y}_{t+i}]^T \),
which is a vector of latitude and longitude at future time $t+i$, for $i = 1, 2, \dots, H$.

\begin{figure}[ht]
    \centering
    \captionsetup{skip=0.5pt}
    \includegraphics[width=0.66\columnwidth]{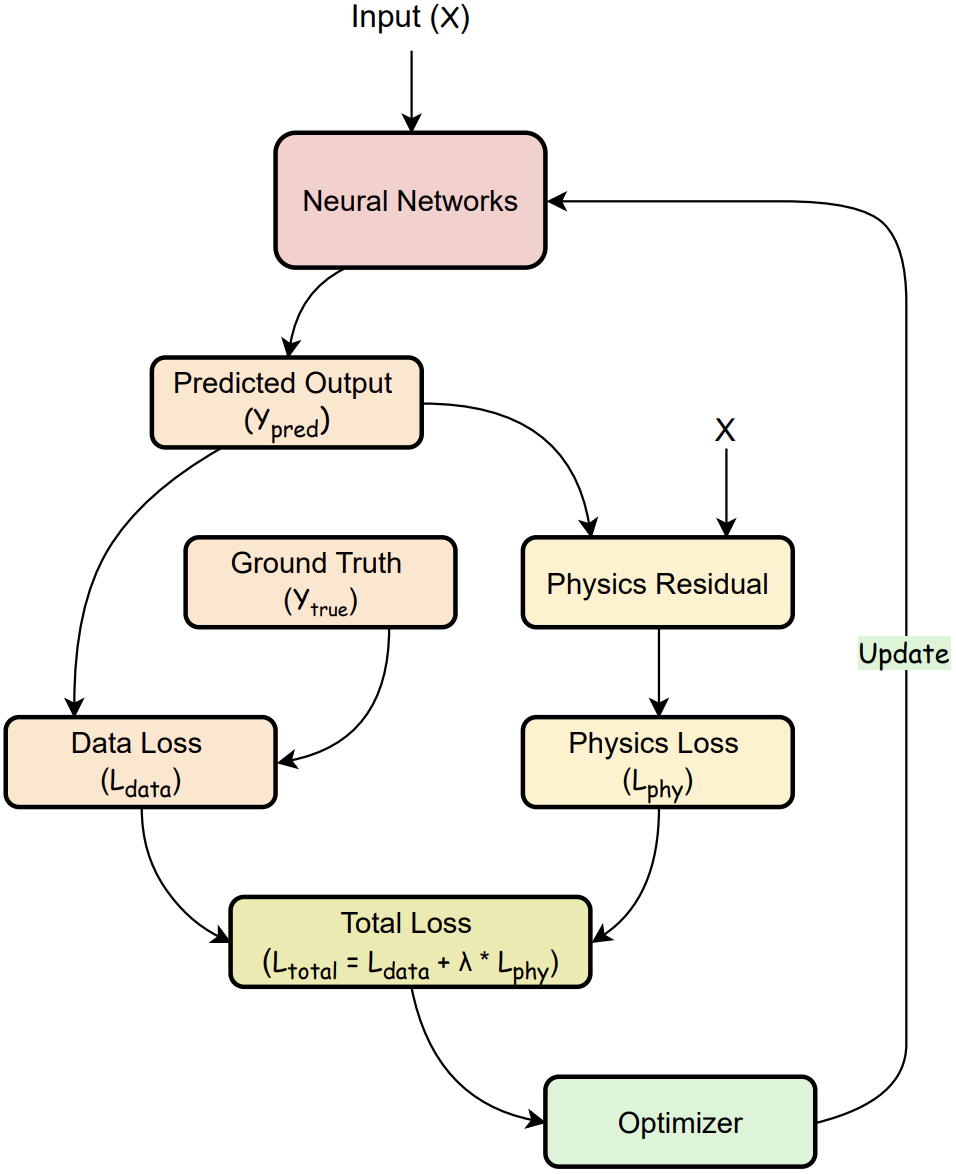}
    \caption{The PINN's training framework optimizes the model by integrating data and physics losses to guide it towards physically informed trajectory predictions. PINN training is an iterative optimization process. In each iteration, the network generates predictions based on input data. A total loss is computed by comparing these predictions to ground truth and physical laws, respectively. The optimizer then minimizes the loss by updating the network's parameters, repeating until convergence.}
    \label{fig:pinn-training}
    \Description{A diagram showing the PINNs training framework, which integrates data loss and physics loss to guide trajectory predictions for vessels.}
\end{figure}

To train the PINN, we minimize a loss function, as depicted in Figure~\ref{fig:pinn-training}. This function is a weighted sum of two distinct terms: the data loss and the physics loss. The \textit{data loss} term ensures the model's predictions align with observed trajectory data, while the \textit{physics loss} term enforces adherence to the underlying kinematic constraints governing vessel motion. This dual objective allows PINNs to complement data-driven approaches by incorporating physical consistency into their predictions.

\subsubsection*{\textbf{Data Loss}}
The data loss component, \( \mathcal{L}_{\text{data}}(\theta) \), quantifies the discrepancy between the predicted trajectory and the ground truth data; to that end, we use the Mean Squared Error (MSE):

\begin{equation}
    \mathcal{L}_{\text{data}}(\theta) = \frac{1}{N} \sum_{i=1}^{N} \sum_{j=1}^{H} \left\| \mathbf{y}_{i,t+j} - \hat{\mathbf{y}}_{i,t+j} \right\|^2
    \label{eq:data_loss}
\end{equation}
where:
\begin{itemize}
    \item \( N \) is the batch size, representing the number of trajectories used in each training iteration.
    \item \( H \) is the prediction horizon.
    \item \( \mathbf{y}_{j, t+i} = [x_{j, t+i}, y_{j, t+i}]^T \) is the ground truth latitude and longitude vector for the \(j\)-th trajectory at time \( t + i \).
    \item \( \hat{\mathbf{y}}_{j, t+i} = f_{\theta}(\mathbf{x}_{j, t}) = [\hat{x}_{j, t+i}, \hat{y}_{j, t+i}]^T \) is the corresponding predicted output vector \( f_{\theta} \), for time step \( i \) for the \( j\)-th trajectory.
    \item \( \| \cdot \|^2 \) denotes the squared Euclidean norm.
\end{itemize}

Minimizing data loss constrains the PINN to approximate a mapping that replicates the observed trajectory patterns in the training data.

\subsubsection*{\textbf{First-Order Physics Loss}}
The physics loss component, \( \mathcal{L}_{\text{phy}}(\theta) \), is the cornerstone of the PINN framework. It enforces physical plausibility by penalizing deviations from a simplified kinematic model of vessel motion. Specifically, we use a \textit{forward Euler approximation} (a first-order finite difference method) of the kinematic equations to estimate the expected change in latitude and longitude over a discrete time step \( \Delta t \). This expected change is then compared against the change predicted by the neural network. To estimate the expected displacement, we consider the input parameters at time \( t \): \( v_t \) (SOG), \( \psi_t \) (COG), \( a_t \) (acceleration), and \( \dot{\psi}_t \) (COG rate). To improve the first-order approximation, we incorporate a midpoint $\psi_{mid, rad}$ derived from the \textit{Taylor series expansion}:
\begin{equation}
    \psi\Bigl(t + \frac{\Delta t}{2}\Bigr) \approx \psi(t) + \dot{\psi}(t) \cdot \frac{\Delta t}{2}
    \label{eq:taylor_cog}
\end{equation}

\noindent Approximating the rate of change of \(\dot{\psi}(t)\) and converting angles to radians, we obtain the midpoint:
\begin{equation}
    \psi_{mid, rad} = \psi_{t, rad} + \frac{1}{2} \cdot\dot{\psi}_{t, rad} \cdot \Delta t
    \label{eq:cog_mid}
\end{equation}
\noindent where \( \psi_{t, rad} = \psi_t \cdot \frac{\pi}{180} \) and \( \dot{\psi}_{t, rad} = \dot{\psi}_t \cdot \frac{\pi}{180} \) are the COG and COG rate in radians, respectively.


Using the midpoint $\psi_{mid, rad}$, the \textit{expected} changes in latitude and longitude, \(\Delta x_{\text{expected}}\) and \(\Delta y_{\text{expected}}\), can be computed using either a small-angle approximation or a great-circle approximation to Earth's curvature; both approaches were adopted in our evaluation. The \textit{small-angle approximation} offers computational efficiency, suitable for short prediction horizons, and is formulated as:
\begin{align}
    \label{eq:delta_lat_approx}
        \Delta x_{expected} &\approx \Bigl(v_t \cdot \cos(\psi_{mid, rad}) \nonumber \\
        &\qquad + \frac{1}{2} \cdot a_t \cdot \Delta t \cdot \cos(\psi_{mid, rad})\Bigr) \cdot \text{factor}  \\
    \label{eq:delta_lon_approx}
        \Delta y_{expected} &\approx \Bigl(v_t \cdot \sin(\psi_{mid, rad}) \nonumber \\
        &\qquad + \frac{1}{2} \cdot a_t \cdot \Delta t \cdot \sin(\psi_{mid, rad})\Bigr) \cdot \frac{\text{factor}}{\cos(x_{t, rad})}
\end{align}
\noindent where \( \text{factor} = \frac{\Delta t}{R} \cdot \frac{180}{\pi} \), \(R\) is Earth's radius, $\Delta x_{expected}$ and $\Delta y_{expected}$ refer to known x and y coordinated displacements, as observed in the data. 

Conversely, for enhanced accuracy, particularly over longer prediction horizons, the \textit{great-circle approximation} is employed, formulated as:
\begin{align}
    \Delta x_{expected} = &\arcsin\left(\sin(x_{t, rad}) \cdot \cos(d) \right. \nonumber \\
    &+ \left. \cos(x_{t, rad}) \cdot \sin(d) \cdot \cos(\psi_{mid, rad})\right) - x_{t,rad} \label{eq:delta_lat_gc} \\
    \Delta y_{expected} = &\arctan2\left(\sin(\psi_{mid, rad}) \cdot \sin(d) \cdot \cos(x_{t, rad}), \right. \nonumber \\
    & \left. \cos(d) - \sin(x_{t, rad}) \cdot \sin(x_{t,rad}) \right) \label{eq:delta_lon_gc}
\end{align}
\noindent where the angular distance \(d\), accounting for Earth's radius \(R\) (approximately 6,371,000 meters), is computed as:
\begin{equation}
    d = \frac{\left(v_t + \frac{1}{2} \cdot a_t \cdot \Delta t\right) \cdot \Delta t}{R}
    \label{eq:angular_distance}
\end{equation}

\noindent The \textit{predicted x and y displacements} in latitude and longitude from the neural network are computed as finite differences:
\begin{align}
    \Delta x_{\text{pred}, t+i} &= \hat{x}_{t+i+1} - \hat{x}_{t+i} 
    \label{eq:delta_x_pred}\\
    \Delta y_{\text{pred}, t+i} &= \hat{y}_{t+i+1} - \hat{y}_{t+i}
    \label{eq:delta_y_pred}
\end{align}

\noindent The mismatch between these predicted and expected changes is then expressed by the \textit{finite difference physics residuals}:
\begin{align}
    r_{x, t+i} &= \Delta x_{\text{pred}, t+i} - \Delta x_{\text{expected}, t+i}
    \label{eq:residual_x}\\
    r_{y, t+i} &= \Delta y_{\text{pred}, t+i} - \Delta y_{\text{expected}, t+i}
    \label{eq:residual_y}
\end{align}

\noindent Finally, the physics loss \( \mathcal{L}_{\text{phy}}(\theta) \) is computed by averaging the sum of squared residuals over all time steps within the prediction horizon and all trajectories in the batch:
\begin{equation}
    \mathcal{L}_{\text{phy}}(\theta) = \frac{1}{N \cdot H} \sum_{i=1}^{N} \sum_{j=1}^{H} \Bigl(r_{x,i, t+j}^2 + r_{y,i, t+j}^2\Bigr)
    \label{eq:physics_loss}
\end{equation}

\subsubsection*{\textbf{Second-Order Physics Loss}}
Despite using midpoint approximations, the physics loss based on the first-order forward Euler method limits its ability to capture dynamics within each time-step (\(\Delta t\)). This limitation leads to discretization errors, especially for longer time steps, as it only considers derivatives at the beginning of the interval. To address this limitation, we introduce a second-order physics loss using Heun's method (a type of Runge-Kutta 2 method), which offers a balance between computational simplicity and improved accuracy over the first-order approach, making it particularly suitable for modeling maritime trajectory dynamics.

Heun's method improves on Euler by incorporating derivative information at both the beginning and an estimated endpoint of the interval through a predictor-corrector approach. First, the \textit{predictor step} uses the forward Euler method to obtain an initial estimate of the state at \(t+\Delta t\), denoted with a superscript \(P\):
\begin{align}
    x_{t+\Delta t}^P &= x_t + \dot{x}_t \cdot \Delta t  \label{eq:heun_pred_x} \\
    y_{t+\Delta t}^P &= y_t + \dot{y}_t \cdot \Delta t  \label{eq:heun_pred_y}
\end{align}

\noindent where \( \dot{x}_t = \frac{dx}{dt}\Bigr|_t \) and \( \dot{y}_t = \frac{dy}{dt}\Bigr|_t \) are the derivatives evaluated using the kinematic model with the vessel's state at time \(t\). 

Second, the \textit{corrector step} refines this estimate by computing the derivatives again, using the kinematic model evaluated at the \textit{predicted} state at \(t+\Delta t\). These derivatives are denoted as follows.
%
\begin{equation*}
    \dot{x}_{t+\Delta t}^P = \left.\frac{dx}{dt}\right|_{x^P_{t+\Delta t}, y^P_{t+\Delta t}}, \quad
    \dot{y}_{t+\Delta t}^P = \left.\frac{dy}{dt}\right|_{x^P_{t+\Delta t}, y^P_{t+\Delta t}}
\end{equation*}

Heun's method then averages the derivatives from the beginning (\(t\)) and the predicted end (\(t+\Delta t\)) to compute the final \textit{expected} change in state over the interval \(\Delta t\):
\begin{align}
    \Delta x_{\text{expected}} &= \frac{1}{2} (\dot{x}_t + \dot{x}_{t+\Delta t}^P) \cdot \Delta t \label{eq:heun_delta_x} \\
    \Delta y_{\text{expected}} &= \frac{1}{2} (\dot{y}_t + \dot{y}_{t+\Delta t}^P) \cdot \Delta t \label{eq:heun_delta_y}
\end{align}

These computed increments, \( \Delta x_{\text{expected}} \) and \( \Delta y_{\text{expected}} \), represent the kinematically \textit{expected changes} for the second-order physics loss \( \mathcal{L}_{\text{phy}}(\theta) \). This loss function then quantifies the discrepancies between these \textit{expected displacements} and the corresponding \textit{predicted displacements} by the neural network, \( \Delta x_{\text{pred}, t+i} \) and \( \Delta y_{\text{pred}, t+i} \) (Eqs.~\ref{eq:delta_x_pred}--\ref{eq:delta_y_pred}). The resulting \textit{residuals}, \(r_{x, t+i}\) and \(r_{y, t+i}\), computed as the difference between predicted and expected displacements (Eqs.~\ref{eq:residual_x}--\ref{eq:residual_y}), are squared and averaged, yielding the final second-order physics loss value (Eq.~\ref{eq:physics_loss}). Minimizing this loss constrains the PINN to produce trajectories that not only fit the data but also strongly adhere to the second-order kinematic approximations provided by Heun's method.

\subsubsection*{\textbf{Total Loss Function}}
The total loss function, \( \mathcal{L}_{\text{total}}(\theta) \), minimized during the PINN training process, is a weighted combination of the data and physics loss:
\begin{equation}
    \mathcal{L}_{\text{total}}(\theta) = \mathcal{L}_{\text{data}}(\theta) + \lambda \cdot \mathcal{L}_{\text{phy}}(\theta)
    \label{eq:total_loss}
\end{equation}
\noindent Here, \( \lambda \geq 0 \) is a non-negative hyperparameter controlling the relative importance of the physics loss in the total loss. By adjusting \( \lambda \), the model can balance its adherence to observed data with its compliance to the kinematic model. 
\\

\noindent For consistent scaling in model training and physical fidelity in loss computation, input kinematic features ($x_t$, $y_t$, $v_t$, $\psi_t$, $a_t$, $\dot{\psi}_t$) are first normalized to $[0,1]$ and then denormalized back to their original units to compute physics-informed residuals. To balance their contribution in the loss, these residuals---quantifying discrepancies between predicted and expected displacements---are normalized by the latitude and longitude ranges in the training set. We further assume that $v_t$, $\psi_t$, $a_t$, and $\dot{\psi}_t$ remain constant at their last observed values throughout the prediction horizon $H$, providing a practical basis for enforcing kinematic constraints without future input.

\section{EVALUATION}
\label{sec:evaluation}
This section details the experimental setup and presents a comprehensive performance analysis of our PINN framework, implemented in Python 3 using the Keras/TensorFlow libraries. 

\subsection{Dataset and Preprocessing}
For evaluation, we selected two distinct AIS datasets from different maritime regions, sourced from AISViz/MERIDIAN~\cite{spadon2024maritime}. The first dataset originates from the Arctic region (LON -95 to -75, LAT 55 to 66), which is characterized by sparse vessel traffic due to a navigation season limited to 4 months annually. In contrast, the second dataset is derived from the Strait of Georgia (LON -128 to -122, LAT 48 to 51), a region known for its high density of vessel traffic, but trajectories with frequent turns and speed alterations. We collected the Arctic dataset, which spans two years (2022-2023) and a dataset of six months (January - June, 2023) from the Strait of Georgia.

For each region, AIS messages were extracted, retaining the following attributes: MMSI, timestamp, latitude, longitude, Speed Over Ground (SOG), Course Over Ground (COG), and ship type. These extracted AIS messages were then filtered based on ship type and vessel identification number (MMSI). Subsequently, vessel trajectories were extracted for each region by grouping the filtered AIS messages according to their MMSI. Finally, for each region, we prepared the AIS dataset for evaluation by applying the following preprocessing steps to each extracted trajectory. 

\subsubsection*{\textbf{Noise Filtering}} Trajectories with invalid MMSI identifiers and AIS messages with duplicate timestamps were removed. To focus on moving vessels, AIS messages with SOG below 0.5 knots (indicating anchored vessels) were excluded. Trajectories were retained only if they exhibited a minimum length of 300 data points. Finally, COG values were wrapped to the range [0, 360) degrees to address issues of non-north GPS settings.

\subsubsection*{\textbf{Time-based Trip Segmentation}} Vessel trajectories, spanning long periods and regions, often include multiple trips and stationary periods. To isolate individual trips, we segmented trajectories if the time gap between consecutive points exceeded 60 minutes.

\subsubsection*{\textbf{Cubic Hermite Interpolation}} 
Trajectory segments are interpolated into 2-minute intervals using \textit{Cubic Hermite splines}. This method uses position and derivative (slope) information at each data point for interpolation, ensuring smooth transitions without overshooting or oscillations and preserving the trajectory's shape. This makes it well-suited for capturing realistic vessel maneuvers, including abrupt direction and speed changes.

\subsubsection*{\textbf{Kinematic Feature Derivation}}
Vessel SOG (\(v_t\)) and COG (\(\psi_t\)) can be influenced by external factors such as wind, ocean currents, and weather conditions. Therefore, deriving acceleration (\(a_t\)) and COG rate (\(\dot{\psi}_t\)) from these values is crucial to capture changes in vessel mobility patterns and modes. Acceleration (\(a_t\)) represents the rate of change of SOG (\(v_t\)) over time, while COG rate (\(\dot{\psi}_t\)) is the rate of change of COG (\(\psi_t\)). In our dataset, SOG values were converted from \(knots\) to \(m/s\). Acceleration (\(m/s^2\)) and COG rate (\(degrees/s\)) were then derived using finite differences as follows:
\begin{equation}
    \begin{aligned}
        a_i &= \frac{v_i - v_{i-1}}{t_i - t_{i-1}}, \quad
        \dot{\psi}_i = \frac{\psi_i - \psi_{i-1}}{t_i - t_{i-1}}
    \end{aligned}
    \label{eq:acceleration_and_cograte}
\end{equation}

\begin{figure}[htbp]
  \centering
  \captionsetup{skip=1pt}
  \includegraphics[width=0.85\columnwidth]{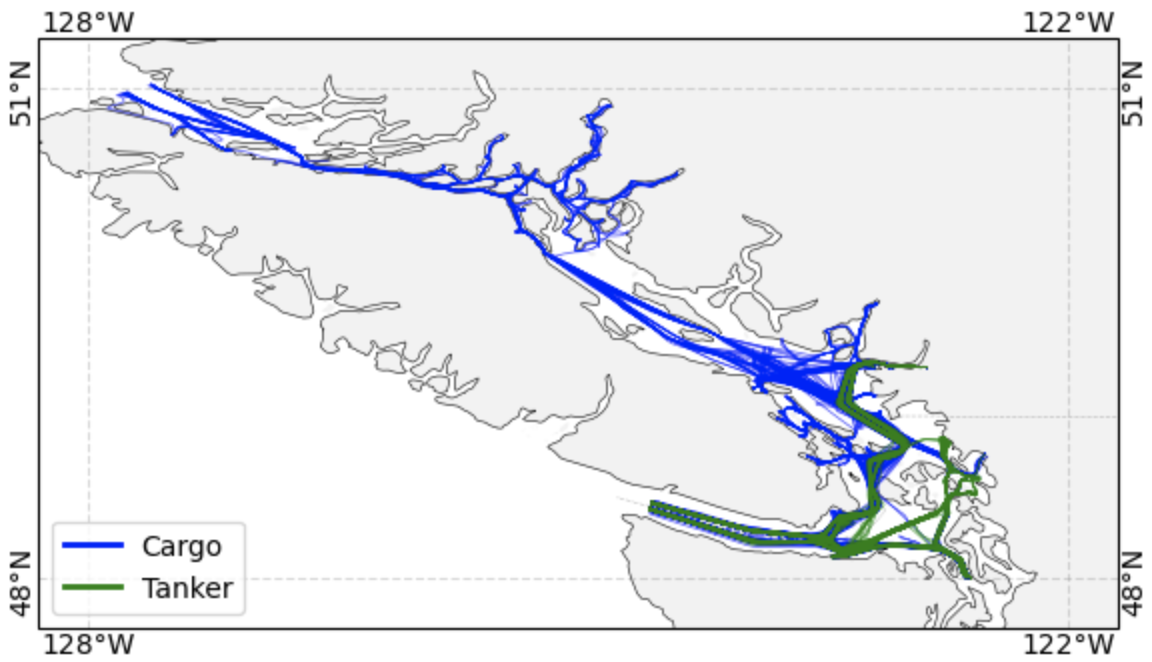}
  \caption{Dense vessel traffic in the Strait of Georgia, depicted by Cargo and Tanker trajectories.}
  \label{fig:map-st-georgia}
\end{figure}

\begin{figure}[htbp]
  \centering
  \captionsetup{skip=1pt}
  \includegraphics[width=0.85\columnwidth]{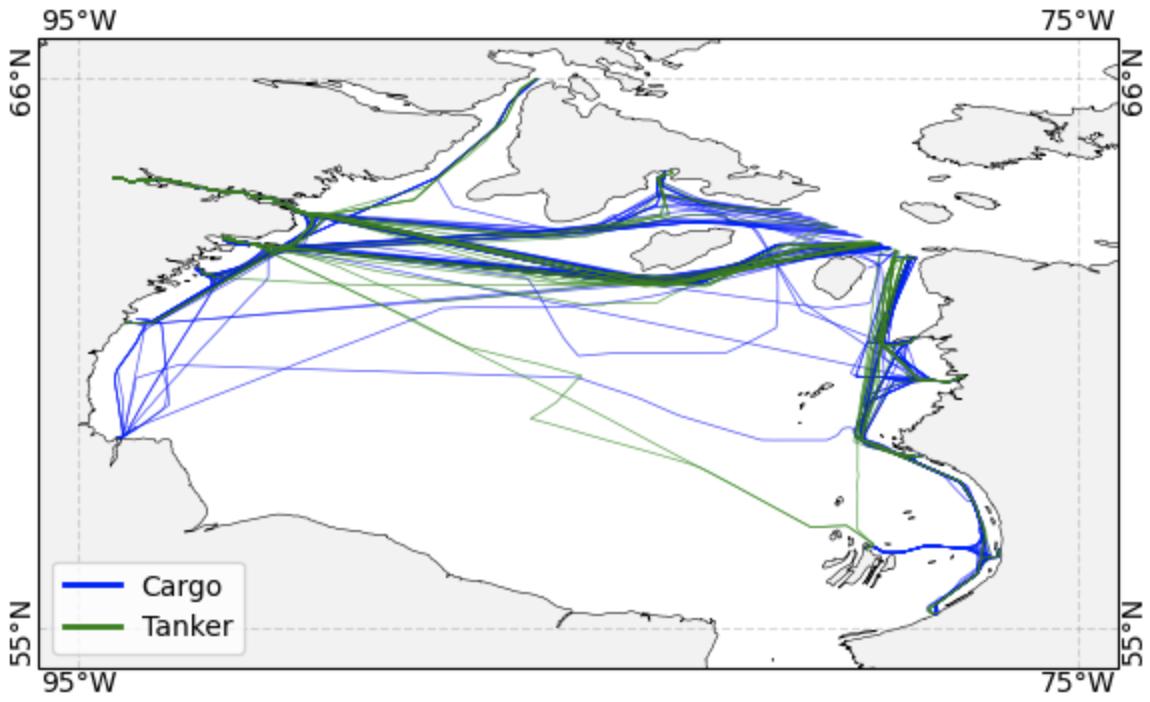}
  \caption{Cargo and Tanker trajectories in the Arctic, highlighting sparse maritime traffic.}
  \label{fig:map-arctic}
\end{figure}

\noindent After preprocessing the trajectories of cargo and tanker vessels from both regions, we kept only $3$ hour segments to ensure consistent input and output sequences for model training and testing. Key statistics of the resulting datasets reveal the following: for the Strait of Georgia, we have $4,315$ cargo segments ($819,265$ data points) and $773$ tanker segments ($161,331$ data points); for the Arctic region, the dataset comprises $277$ cargo segments ($143,399$ data points) and $216$ tanker segments ($76,675$ data points). Their spatial distributions are visualized in Figures~\ref{fig:map-st-georgia} and \ref{fig:map-arctic}, respectively.

\vspace{-3pt}

\subsection{Evaluation Metrics}
To evaluate the accuracy of the prediction models, we employed both \textit{classical regression} metrics, such as Mean Absolute Error (MAE) and Mean Squared Error (MSE), and \textit{state-of-the-art} trajectory prediction metrics, such as Average Displacement Error (ADE) and Final Displacement Error (FDE). ADE and FDE directly quantify spatial accuracy, which is critical for navigational decision making in maritime contexts. 

MSE averages the squared differences between predicted (\(\hat{y}_i\)) and actual (\(y_i\)) locations. Although MSE is computationally straightforward and emphasizes larger errors, it is highly sensitive to outliers. MAE averages the absolute differences, treating all errors equally and exhibiting greater robustness to outliers. The formulas for MAE and MSE are given in Eq.~\ref{ch2_eq:reg-metrics}, where \(n\) denotes the sample size.

\begin{equation}
    \label{ch2_eq:reg-metrics}
    \text{MAE} = \frac{1}{n} \sum_{i=1}^{n} |y_i - \hat{y}_i|, \quad
    \text{MSE} = \frac{1}{n} \sum_{i=1}^{n} (y_i - \hat{y}_i)^2
\end{equation}

ADE represents the average distance between predicted \((\hat{x}_i, \hat{y}_i)\) and actual \((x_i, y_i)\) locations over all time steps, providing a comprehensive measure of trajectory accuracy. Lower ADE values indicate better overall prediction. FDE, in contrast, measures only the distance between the final predicted \((\hat{x}_n, \hat{y}_n)\) and actual \((x_n, y_n)\) locations. Lower FDE values suggest better accuracy in predicting the trajectory endpoint and reduced error accumulation over time. The formulas for ADE and FDE, using \textit{Haversine} distance \(H_d\), are given in Eq.~\ref{ch2_eq:dist-metrics}, where \(n\) is the trajectory length.
\begin{equation}
    \label{ch2_eq:dist-metrics}
    \begin{aligned}
    \text{ADE}  &= \frac{1}{n} \sum_{i=1}^{n} H_d\bigl((x_i, y_i), (\hat{y}_i, \hat{y}_i)\bigr) \\
    \text{FDE}  &= H_d\bigl((x_n, y_n), (\hat{x}_n, \hat{y}_n)\bigr)
    \end{aligned}
\end{equation}
where \(H_d\) represents the \textit{Haversine} distance between two points, calculated as:
\begin{equation}
    \label{ch2_eq:hversine}
    H_{d} = 2 \cdot R \cdot \arcsin \sqrt{\sin^{2} \left(\frac{\Delta x}{2}\right) + \cos (x_1) \cdot \cos (x_2) \cdot \sin ^ {2} \left(\frac{\Delta y} {2}\right)}
\end{equation}
\noindent Here, \(R\) represents the Earth's radius, and \(\Delta x = x_2 - x_1\), \(\Delta y = y_2 - y_1\), where \(x\) and \(y\) denote latitude and longitude in radians, respectively. Distance errors are computed in meters in this evaluation.

\subsection{Setting Model I/O and Hyperparameters}

\subsubsection*{\textbf{DL Model Architectures}}
We selected various state-of-the-art Deep Learning (DL) models that have been widely used by researchers in trajectory prediction and sequence modeling. This selection consists of a range of architectures, including Recurrent Neural Network (RNN)-based models (LSTMs and GRUs), Convolutional Neural Networks (CNNs), Temporal Convolutional Networks (TCNs), Hybrid architectures (ConvLSTMs), and Transformers. 

\subsubsection*{\textbf{Data Preparation for Model I/O}}
To prepare the dataset for DL models, we begin by performing feature-wise normalization using Min-Max scaling and subsequently apply a sliding-window approach. For each trajectory, we extracted the input sequences \(X\) and the corresponding target sequences \(Y\) by sliding a window of length \(W_{\text{in}} + W_{\text{out}}\). Given a trajectory and denoting \(t\) as the last time step of the input window, the input sequence \(X\) comprises \(W_{\text{in}}\) consecutive feature vectors ending at time step \(t\):
\begin{equation*}
    X_t = \{\mathbf{x}_{t - W_{\text{in}} + 1}, \mathbf{x}_{t - W_{\text{in}} + 2}, \dots, \mathbf{x}_{t}\}
\end{equation*}

\noindent The corresponding target sequence \(Y\), representing the prediction horizon \(W_{\text{out}}\), consists of \(W_{\text{out}}\) output vectors starting from time step \(t+1\):
\begin{equation*}
    Y_t = \{\mathbf{y}_{t+1}, \mathbf{y}_{t+2}, \dots, \mathbf{y}_{t + W_{\text{out}}}\}
\end{equation*}

\noindent Applying this sliding window generated input-target pairs $(X_t, Y_t)$ from all trajectories for training, validation, and testing. For each vessel type, $10\%$ of the trajectories were assigned to the test set. Of the remaining trajectories, $80\%$ for training and $20\%$ for validation.

\subsubsection*{\textbf{Hyperparameter Settings}}
The physics loss weight, \(\lambda\), is a crucial hyperparameter for PINNs. To assess its influence, we performed experiments on \(\lambda\) values ranging from \(0.0001\) to \(1.0\). All other hyperparameters were kept consistent across models and datasets. All models were trained with Adam optimizer (initial learning rate $0.001$) for up to $50$ epochs and with a batch size of $32$. The models were optimized for MSE, but performance was monitored using both MSE and MAE. Hidden layers used ReLU activations, while the output layers employed linear activations. Training was regularized using Early Stopping and the ReduceLROnPlateau scheduler, based on validation performance. 

%
We used an encoder-only Transformer configured with $2$ layers, $128$ model dimensions, $4$ attention heads, and $256$ feed-forward dimensions, along with dropout and default positional encoding. For TCN, we used residual blocks with increasing dilation rates $\{2^i\}_{i=0}^{4}$. For other models, we used $64$ hidden units for 1-layer and $64/32$ units for 2-layer setups. 

\subsection{Performance Analysis}
\label{sub:performance-analysis}
To comprehensively assess efficacy, we evaluated our PINN-integrated models against baselines without physical constraints, based on the following criteria:

\begin{enumerate}
    \item \textbf{Model Complexity:} We varied the complexity of DL models to investigate their ability to balance data-driven accuracy with physics constraints. \textbf{Basic} refers to simpler configurations (\textit{e.g.}, single-layer LSTMs, GRUs, CNNs, or fewer TCN residual blocks), while \textbf{Complex} refers to models with deeper or wider layers, or additional residual blocks. 

    \item \textbf{Waterway Complexity:} We considered waterway complexity in the performance analysis, as areas such as ports or waypoints are more complex than open seas. Two cases were considered: \textbf{Case 1} (all sliding windows) includes both dynamic and open-seas, whereas \textbf{Case 2} (beginning, middle, and end windows) focuses on dynamic areas.
    
    \item \textbf{Prediction Horizon:} We varied the lengths of the prediction horizon to assess the models' robustness against temporal uncertainty and error accumulation.

    \item \textbf{Score Selection:} Instead of relying on the lowest ADE/FDE scores for assessment, we selected these scores based on physics plausibility, robust validation performance, and synergistic convergence, as monitored through training/validation data loss and physics loss curves across epochs.

    \item \textbf{Numerical Approximation:} We assessed the performance based on the order of numerical approximations. 
\end{enumerate}

\subsubsection*{\textbf{Arctic Dataset}} Table~\ref{tab:result-a1} presents the mean ADE and FDE for various DL models, evaluated with and without PINN regularization across two levels of model complexity. The results consistently show that incorporating PINN reduces both ADE (up to $32\%$) and FDE (up to $27\%$) across nearly all model types and complexity levels. However, as model complexity increases, the performance gains for architectures like ConvLSTM and TCN do not improve significantly, likely due to their higher data requirements for optimal convergence. Nonetheless, these complex models with PINN still achieve significantly lower errors compared to their non-PINN counterparts and are close to their basic implementations, suggesting that PINN regularization enables effective model training even with relatively small datasets.

\begin{table}[htbp]
  \centering
  \captionsetup{skip=0.5pt}
  \caption{Model performance by level of model complexity.}
  \label{tab:result-a1}
  \resizebox{\columnwidth}{!}{%
  \begin{tabular}{llcccc}
    \toprule
    \multicolumn{6}{c}{\(Dataset = Arctic\), \(Vessel = Tanker\), \(W_{in}=30min\), \(W_{out}=30min\), \(\Delta t = 2min\)} \\
    \midrule
    \multicolumn{2}{c}{\multirow{2}{*}{\textbf{Models}}} & \multicolumn{2}{c}{\textbf{No PINN}} & \multicolumn{2}{c}{\textbf{First--Order PINN}} \\
    \cmidrule(lr){3-4} \cmidrule(lr){5-6}
    \multicolumn{2}{c}{} & \textbf{ADE} & \textbf{FDE} & \textbf{ADE} & \textbf{FDE} \\
    \midrule
    \multirow{5}{*}{\rotatebox[origin=c]{90}{\textbf{Basic}}}
    & LSTM & 2584 $\pm$ 840 & 4354 $\pm$ 1508 & \textbf{2296} $\pm$ 679 & 3950 $\pm$ 1469 \\
    & GRU  & 2769 $\pm$ 793 & 4913 $\pm$ 1383 & \textbf{2477} $\pm$ 741 & 4447 $\pm$ 1288 \\
    & CNN  & 3385 $\pm$ 1036 & 5409 $\pm$ 1818 & \textbf{3336} $\pm$ 841 & 5406 $\pm$ 1300 \\
    & ConvLSTM  & 1613 $\pm$ 529 & 2615 $\pm$ 873 & \textbf{1384} $\pm$ 454 & 2215 $\pm$ 871 \\
    & TCN  & 2036 $\pm$ 751 & \textbf{3428} $\pm$ 1386 & \textbf{1811} $\pm$ 643 & 3444 $\pm$ 1334 \\
    \midrule
    \multirow{5}{*}{\rotatebox[origin=c]{90}{\textbf{Complex}}}
    & LSTM & 2176 $\pm$ 789 & 3483 $\pm$ 1532 & \textbf{2081} $\pm$ 655 & 3474 $\pm$ 1322 \\
    & GRU  & 2542 $\pm$ 833 & 4330 $\pm$ 1288 & \textbf{2218} $\pm$ 635 & 3898 $\pm$ 1430 \\
    & CNN  & 2687 $\pm$ 939 & 4570 $\pm$ 1530 & \textbf{2552} $\pm$ 698 & 4285 $\pm$ 1262 \\
    & ConvLSTM  & 1961 $\pm$ 737 & 3010 $\pm$ 1432 & \textbf{1596} $\pm$ 503 & 2383 $\pm$ 872 \\
    & TCN  & 2652 $\pm$ 1703 & 4479 $\pm$ 2367 & \textbf{1800} $\pm$ 520 & 3232 $\pm$ 1085 \\
    \bottomrule
  \end{tabular}
  } %
\end{table}

Table~\ref{tab:result-a2} shows the model performance trends as prediction horizons increase based on best performing models, ConvLSTM and TCN. It is evident that the First--Order PINN consistently reduces ADE across all prediction horizons. The percentage decrease compared to No PINN models ranges from approximately $11\%$ to over $37\%$ for \textit{basic} models, while it ranges from $15\%$ to $33\%$ for \textit{complex} models. Overall, the percentage reduction in ADE through the integration of PINN is consistently positive on all horizons and models, demonstrating its robust performance enhancement.

\begin{table}[htbp]
  \centering
  \captionsetup{skip=0.5pt}
  \caption{Model performance by prediction horizon length.}
  \label{tab:result-a2}
  \resizebox{\columnwidth}{!}{%
  \footnotesize
  \begin{tabular}{llcccccc}
    \toprule
    \multicolumn{8}{c}{\footnotesize Dataset = Arctic, Vessel = Tanker, \(W_{in}=30\text{m}\), \(W_{out}=10/20/30\text{m}\), \(\Delta t=2\text{min}\)} \\
    \midrule
    & \multirow{2}{*}{\textbf{Models}} & \multicolumn{3}{c}{\textbf{No PINN (ADE)}} & \multicolumn{3}{c}{\textbf{First Order PINN (ADE)}} \\
    \cmidrule(lr){3-5} \cmidrule(lr){6-8}
    & & \textbf{10m} & \textbf{20m} & \textbf{30m} & \textbf{10m} & \textbf{20m} & \textbf{30m} \\
    \midrule
    \multirow{2}{*}{\textbf{Basic}}
    & ConvLSTM & 1047 & 1258 & 1613 & 895 & 1091 & 1384 \\
    & TCN      & 1259 & 1565 & 2036 & 788 & 1254 & 1811 \\
    \midrule
    \multirow{2}{*}{\textbf{Complex}}
    & ConvLSTM & 1395 & 1616 & 1961 & 1189 & 1318 & 1596 \\
    & TCN      & 1384 & 1997 & 2652 & 1018 & 1330 & 1800 \\
    \bottomrule
  \end{tabular}
  }
\end{table}

As stated before, vessel maneuvers differ significantly between dynamic areas (\textit{e.g.}, ports and waypoints), while course and speed adjustments are less frequent in open seas. Therefore, averaging performance across entire trajectories (\textit{Case 1}) can obscure the nuanced benefits of physics-informed models, particularly in areas like the Arctic (see Figure~\ref{fig:map-arctic}). To specifically assess the impact of physics approximations, we also strategically selected test windows (\textit{Case 2}) to capture varying navigation contexts. Table~\ref{tab:result-a3} reveals that prediction errors (ADE and FDE) are higher in Case 2 than in Case 1, indicating greater complexity in the selected test windows. Notably, the integration of First--Order PINN consistently reduces errors in both cases, highlighting its effectiveness with smaller datasets regardless of navigational complexity.

\begin{table}[htbp]
  \centering
  \captionsetup{skip=0.5pt}
  \caption{Model performance by complexity of water area.}
  \label{tab:result-a3}
  \resizebox{\columnwidth}{!}{%
  \footnotesize
  \begin{tabular}{llcccc} 
    \toprule
    \multicolumn{6}{c}{\footnotesize Dataset = Arctic, Vessel = Tanker, \(W_{in}=30\text{m}\), \(W_{out}=30\text{m}\), \(\Delta t=2\text{min}\)} \\
    \midrule
    & \multirow{2}{*}{\textbf{Basic Models}} & \multicolumn{2}{c}{\textbf{No PINN}} & \multicolumn{2}{c}{\textbf{First-Order PINN}} \\
    \cmidrule(lr){3-4} \cmidrule(lr){5-6}
    & &  \textbf{ADE} & \textbf{FDE} & \textbf{ADE} & \textbf{FDE} \\ 
    \midrule
    \multirow{2}{*}{\textbf{Case 1}}
    & ConvLSTM & 1613 $\pm$ 529 & 2615 $\pm$ 873  & 1384 $\pm$ 454 & 2215 $\pm$ 871  \\
    & TCN      & 2036 $\pm$ 751 & \textbf{3428} $\pm$ 1386 & 1811 $\pm$ 643 & 3444 $\pm$ 1334 \\
    \midrule
    \multirow{2}{*}{\textbf{Case 2}}
    & ConvLSTM & 2169 $\pm$ 776 & 3248 $\pm$ 1202 & 1716 $\pm$ 596 & 2674 $\pm$ 1035 \\ 
    & TCN      & 2409 $\pm$ 861 & 4057 $\pm$ 1556 & 1999 $\pm$ 886 & 3512 $\pm$ 1734 \\ 
    \bottomrule
  \end{tabular}
  }
\end{table}

\subsubsection*{\textbf{Strait of Georgia Dataset}} As a dense and narrow waterway, the Georgia Strait exhibits more complex and dynamic vessel movement patterns compared to the Arctic. Although it is not a large dataset, the $6$-month Georgia dataset contains more trajectories than $2$ years of Arctic data. As shown in Table~\ref{tab:result-g1}, overall average and final displacement errors were significantly reduced compared to the Arctic dataset (see Table~\ref{tab:result-g1}) as models converged better with more data. Unlike the Arctic, both open seas and port areas are geometrically complex in this dataset. Consequently, we computed errors for all sliding windows (Case 1), where the PINN approach yielded ADE error reductions of up to approximately $12\%$ and FDE $15\%$, highlighting its subtle yet crucial benefits.

\begin{table}[htbp]
  \centering
  \captionsetup{skip=0.5pt}
  \caption{Displacement errors - No PINN \textit{vs.} First-Order PINN}
  \label{tab:result-g1}
  \resizebox{\columnwidth}{!}{%
  \begin{tabular}{lcccc}
    \toprule
    \multicolumn{5}{c}{\(Dataset = Georgia\), \(Vessel = Tanker\), \(W_{in}=30min\), \(W_{out}=30min\), \(\Delta t = 2min\)} \\
    \midrule
    \multirow{2}{*}{\textbf{Complex Models}} & \multicolumn{2}{c}{\textbf{No PINN}} & \multicolumn{2}{c}{\textbf{First--Order PINN}} \\
    \cmidrule(lr){2-3} \cmidrule(lr){4-5}
    & \textbf{ADE}  & \textbf{FDE}   & \textbf{ADE} & \textbf{FDE} \\
    \midrule
    LSTM      & 740 $\pm$ 219  & 1394 $\pm$ 515 & \textbf{722} $\pm$ 204 & 1358 $\pm$ 461 \\
    GRU       & 805 $\pm$ 222  & 1566 $\pm$ 529 & \textbf{799} $\pm$ 224 & 1549 $\pm$ 500 \\
    CNN       & 1318 $\pm$ 480 & 2288  $\pm$ 838 & \textbf{1162} $\pm$ 378 & 2014 $\pm$ 682 \\
    ConvLSTM  & \textbf{637} $\pm$ 197  & 1177 $\pm$ 447 & 654 $\pm$ 203 & 1237  $\pm$ 427 \\
    TCN       & 800 $\pm$ 233  & 1596 $\pm$ 518 & \textbf{784} $\pm$ 216 & 1369 $\pm$ 494 \\
    \bottomrule
  \end{tabular}
  }
\end{table}

Furthermore, MAE and MSE were computed for both latitude and longitude (see Table~\ref{tab:result-g2}), which show that longitude errors are consistently larger than latitude errors, indicating that east-west motion is harder to predict. The first-order PINN significantly reduced longitude errors (MAE by $2$--$10\%$ and MSE by up to $23\%$), especially for the CNN model. This indicates that the kinematic constraint mainly reduces the noisier longitude errors, whereas the already small latitude errors remain largely unchanged and contribute little to the overall prediction.

\begin{table}[htbp]
  \centering
  \captionsetup{skip=0.5pt}
  \caption{PINN Impact on Spatial Dimensions.}
  \label{tab:result-g2}
  \resizebox{\columnwidth}{!}{%
  \begin{tabular}{lcccccccc} %
    \toprule
    \multicolumn{9}{c}{\(Dataset = Georgia\), \(Vessel = Tanker\), \(W_{in}=30min\), \(W_{out}=30min\), \(\Delta t = 2min\)} \\ 
    \midrule
    \multirow{3}{*}{\makecell{\textbf{Complex}\\\textbf{Models}}} 
    & \multicolumn{4}{c}{\textbf{No PINN}} & \multicolumn{4}{c}{\textbf{First--Order PINN}} \\ 
    \cmidrule(lr){2-5} \cmidrule(lr){6-9} 
    & \multicolumn{2}{c}{\textbf{MAE}} & \multicolumn{2}{c}{\textbf{MSE}} 
    & \multicolumn{2}{c}{\textbf{MAE}} & \multicolumn{2}{c}{\textbf{MSE}} \\ 
    \cmidrule(lr){2-3} \cmidrule(lr){4-5} \cmidrule(lr){6-7} \cmidrule(lr){8-9} 
    & LAT & LON & LAT & LON & LAT & LON & LAT & LON \\ 
    \midrule
    LSTM   & 0.0041 & 0.0066 & 0.00005 & 0.00012 & 0.0041 & \textbf{0.0064} & 0.00005 & 0.00012 \\ 
    GRU    & 0.0046 & 0.0070 & 0.00006 & 0.00014 & 0.0046 & \textbf{0.0069} & 0.00006 & 0.00014 \\ 
    CNN   & 0.0074 & 0.0115 & 0.00016 & 0.00035 & 0.0065 & \textbf{0.0104} & 0.00011 & 0.00027 \\ 
    ConvLSTM & 0.0036 & \textbf{0.0057} & 0.00004 & 0.00010 & 0.0036 & 0.0059 & 0.00004 & 0.00010 \\ 
    TCN    & 0.0044 & 0.0072 & 0.00006 & 0.00016 & 0.0043 & \textbf{0.0070} & 0.00006 & 0.00015 \\ 
    \bottomrule
  \end{tabular}
  }
\end{table}

\subsubsection*{\textbf{First--Order \textit{vs.} Second--Order Physics}}
We also analyzed whether adding second-order dynamics to PINN improves predictive accuracy compared to a first-order approximation, specifically on the smaller Arctic dataset. As shown in Table~\ref{tab:result-a4}, switching from a first-order to a second-order PINN yields marginal gains in ADE and FDE across models (\textit{e.g.}, ConvLSTM reduced ADE slightly from $1596$ to $1558$). This subtle difference is expected, as the Arctic dataset reflects less dynamic conditions, featuring predominantly open-sea navigation with low traffic density. Also, predicting over a short horizon ($\leq 30$~min) with small time steps ($2$~min) means higher-order effects have limited time to accumulate and significantly impact the predictions. Moreover, the first-order PINN has already achieved a significant improvement. For TCN, the second-order PINN does not yield further improvement, but when considering the complex area (Case 2), it reduces ADE and FDE to $1927$ and $3408$ respectively, from first-order errors (see Table~\ref{tab:result-a3}).

\begin{table}[htbp!]
  \centering
  \captionsetup{skip=0.5pt}
  \caption{Model performance by order of approximations.}
  \label{tab:result-a4}
  \resizebox{\columnwidth}{!}{%
  \begin{tabular}{lcccc}
    \toprule
    \multicolumn{5}{c}{\(Dataset = Arctic\), \(Vessel = Tanker\), \(W_{in}=30min\), \(W_{out}=30min\), \(\Delta t = 2min\)} \\
    \midrule
    \multirow{2}{*}{\textbf{Complex Models}} & \multicolumn{2}{c}{\textbf{First--Order PINN}} & \multicolumn{2}{c}{\textbf{Second--Order PINN}} \\
    \cmidrule(lr){2-3} \cmidrule(lr){4-5}
    & \textbf{ADE}  & \textbf{FDE}   & \textbf{ADE} & \textbf{FDE} \\
    \midrule
    LSTM     & 2081 $\pm$ 655 & 3474 $\pm$ 1322 & \textbf{1973} $\pm$ 507 & \textbf{3408} $\pm$ 1252 \\
    GRU      & 2218 $\pm$ 635 & 3898 $\pm$ 1430 & \textbf{2198} $\pm$ 587 & \textbf{3841} $\pm$ 1379 \\
    CNN      & 2552 $\pm$ 698 & 4285 $\pm$ 1262 & \textbf{2222} $\pm$ 715 & \textbf{3750} $\pm$ 1346 \\
    ConvLSTM & 1596 $\pm$ 503 & 2383 $\pm$ 872  & \textbf{1558} $\pm$ 536 & \textbf{2319} $\pm$ 900  \\
    TCN      & \textbf{1800} $\pm$ 520 & \textbf{3232} $\pm$ 1085 & 1823 $\pm$ 610 & 3465 $\pm$ 1332 \\
    \bottomrule
  \end{tabular}
  }
\end{table}

\subsubsection*{\textbf{Transformer}} To further assess PINN efficacy, we also evaluated an encoder-only Transformer, whose capacity to model long-range dependencies suits complex sequential tasks like trajectory prediction. Given the Transformer's inherent requirement for substantial training data to achieve convergence, we conducted experiments using the Georgia dataset. As presented in Table~\ref{tab:result-g3}, both ADE and FDE errors exhibited reductions when switching from No PINN settings to First-Order PINN in both cases.

\begin{table}[htbp]
  \centering
  \captionsetup{skip=0.5pt}
  \caption{Model performance by complexity of water area and order of numerical approximations.}
  \label{tab:result-g3}
  \resizebox{\columnwidth}{!}{%
  \begin{tabular}{ccccccc}
    \toprule
    \multicolumn{7}{c}{\(Dataset = Georgia\), \(Vessel = Tanker\), \(W_{in}=30min\), \(W_{out}=30min\), \(\Delta t = 2min\)} \\
    \midrule
    \multirow{2}{*}{\textbf{Transformer}} & \multicolumn{2}{c}{\textbf{No PINN}} & \multicolumn{2}{c}{\textbf{First--Order PINN}} & \multicolumn{2}{c}{\textbf{Second--Order PINN}} \\
    \cmidrule(lr){2-3} \cmidrule(lr){4-5} \cmidrule(lr){6-7}
    & \textbf{ADE} & \textbf{FDE} & \textbf{ADE} & \textbf{FDE} & \textbf{ADE} & \textbf{FDE} \\
    \midrule
    \textbf{Case 1} & 605 & 1046 & \textbf{554} & \textbf{992}  & 573 & 1023 \\
    \textbf{Case 2} & 691 & 1112 & \textbf{629} & \textbf{1087} & 645 & 1092 \\
    \bottomrule
  \end{tabular}
  }
\end{table}

\begin{figure}[htbp]
  \centering
  \captionsetup{skip=0.5pt, justification=centering}
  \includegraphics[width=.95\columnwidth]{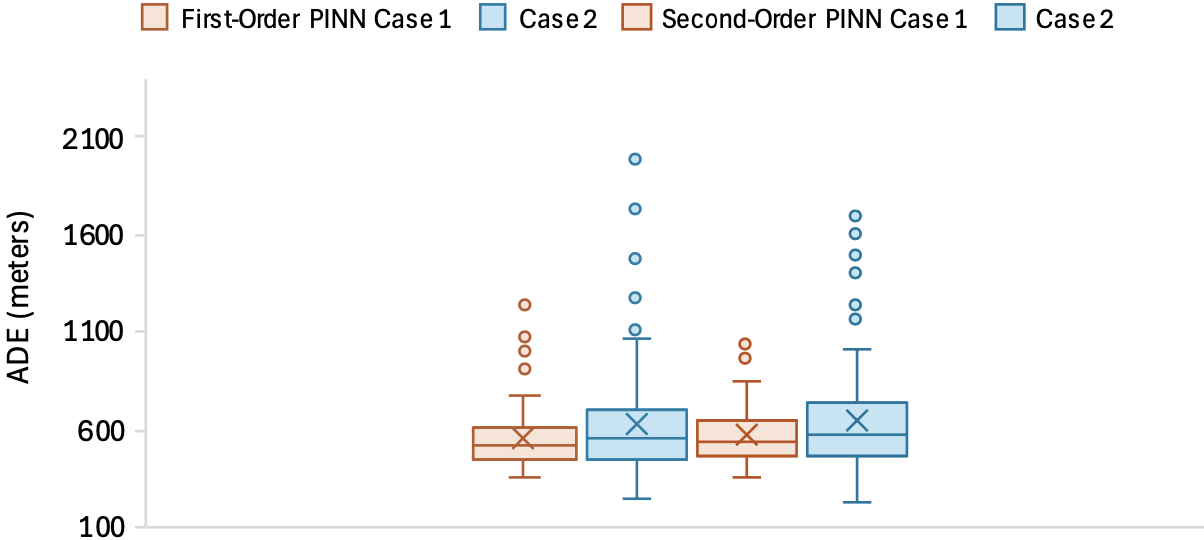}
  \caption{Transformer - First Order \textit{versus} Second Order.}
  \label{fig:transformer-g1}
\end{figure}

Although Second--order PINN did not yield further improvements in terms of ADE and FDE magnitudes, it is noteworthy that the standard deviation of displacement errors decreased, as evidenced by the reduced spread in the box plot of displacement errors shown in Figure~\ref{fig:transformer-g1}.

\subsubsection*{\textbf{PINNs Complexity}} 
\begin{figure}[htbp]
  \centering
  \captionsetup{skip=0.3pt, justification=centering}
  \includegraphics[width=.8\columnwidth]{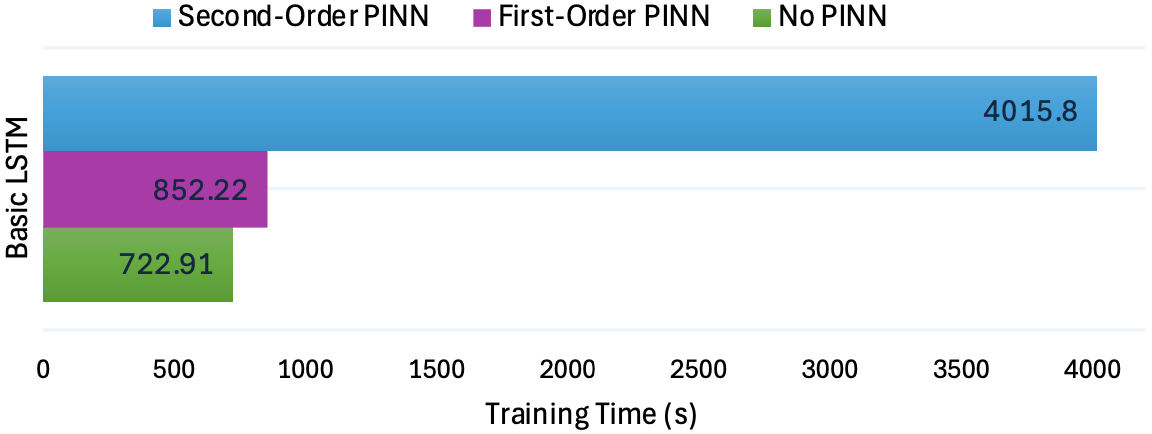}
  \caption{LSTM training time comparison (Arctic Dataset).}
  \label{fig:pinn-time}
\end{figure}

PINN training typically requires more time than standard supervised learning due to the additional physics loss computation. As shown in Figure~\ref{fig:pinn-time}, using our finite difference approach, the training time increases moderately for the first-order PINN and significantly for the second-order PINN compared to the non-PINN baseline. Moreover, tuning the weight parameter \(\lambda\) is necessary to achieve optimal convergence. However, the proposed finite difference approach is faster and consumes less memory compared to standard PINNs that rely on automatic differentiation, which requires storing and computing intermediate operations and their derivatives via a computational graph.

\subsubsection*{\textbf{Sample Prediction Results}} 
\begin{figure}[htbp]
  \centering
  \captionsetup{skip=0.5pt, justification=centering}
  \includegraphics[width=\columnwidth]{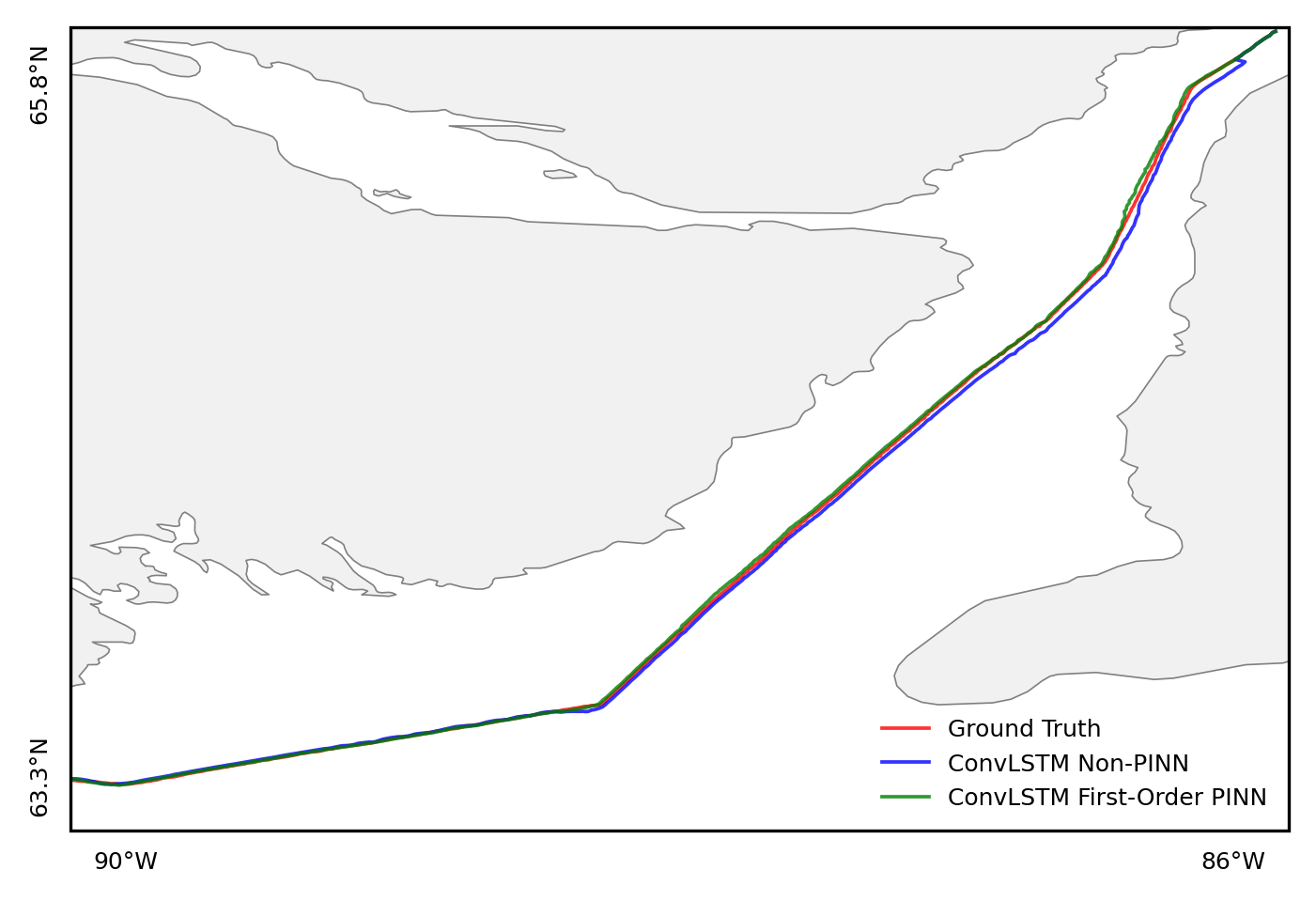}
  \caption{Arctic - Observed \textit{vs.} ConvLSTM Predicted Tracks.}
  \label{fig:arctic-viz}
\end{figure}

\begin{figure}[htbp]
  \centering
  \captionsetup{skip=1pt, justification=centering}
  \includegraphics[width=\columnwidth]{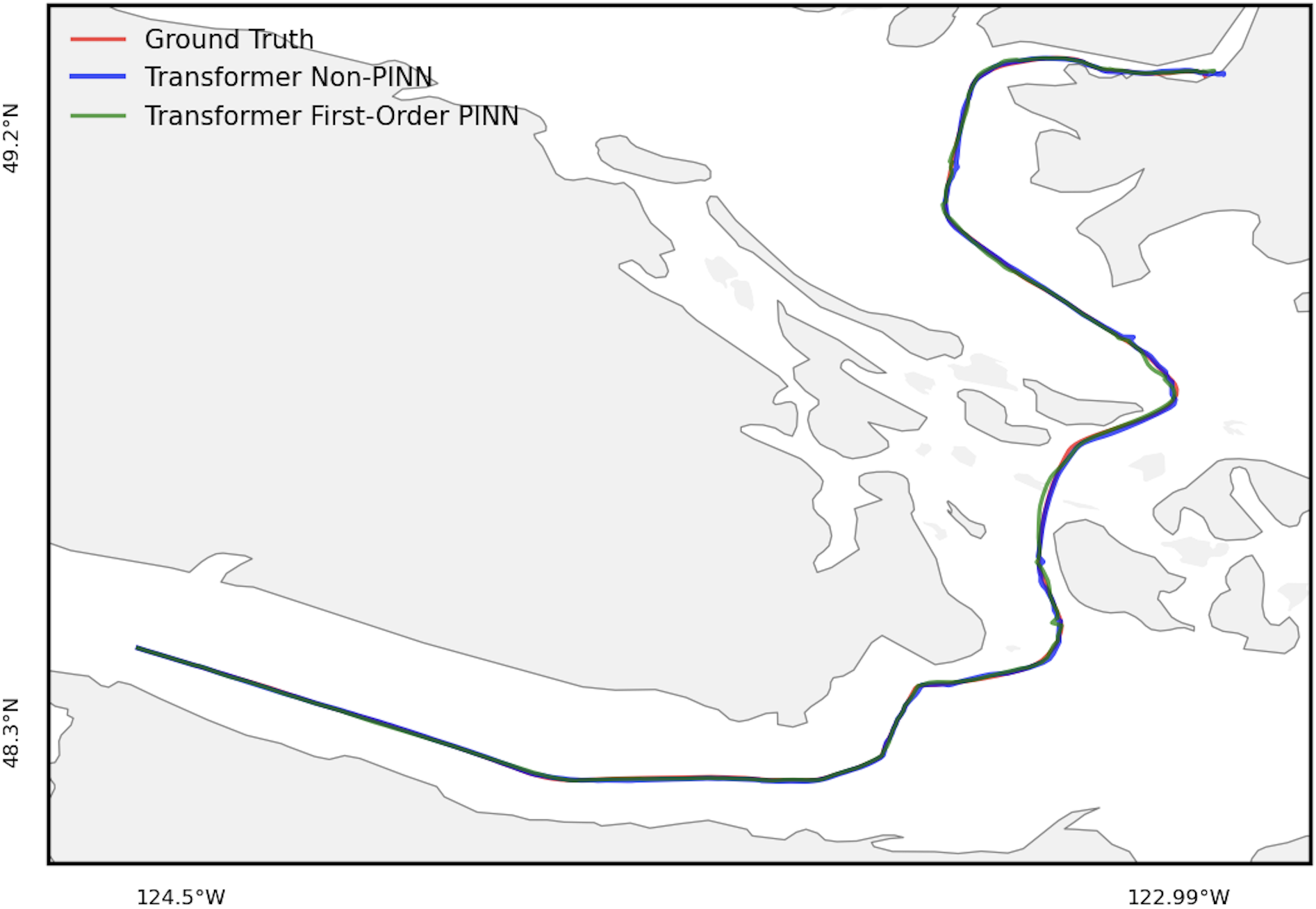}
  \caption{Georgia - Observed \textit{vs.} Transformer Tracks.}
  \label{fig:georgia-viz}
\end{figure}

Figures~\ref{fig:arctic-viz} and~\ref{fig:georgia-viz} show predicted tracks from the Arctic and Georgia regions, respectively. These tracks are generated by combining predictions across all sliding windows using the best performing models for each region: ConvLSTM for the Arctic and Transformer for Georgia. As observed in Figure~\ref{fig:arctic-viz}, the track predicted by the Non-PINN ConvLSTM significantly deviates from the ground truth. The ADE of tracks predicted by ConvLSTM is 1417m for Non-PINN and 982m for First-Order PINN. In Figure~\ref{fig:georgia-viz}, the track predicted by the Non-PINN Transformer shows many spikes, indicating notable deviations from the reference line. The ADE in this case 535m and 463m for Non-PINN and First-Order PINN, respectively. This also proves that even when a comparatively good amount of data is available, a PINN model can enforce kinematic constraints and improve prediction accuracy.

\section{CONCLUSION AND FUTURE WORK}
\label{sec:conclusion}
Recent DL-based approaches for vessel trajectory prediction are data-intensive and lack explicit integration of vessel motion dynamics. Consequently, their prediction accuracy suffers due to sudden vessel maneuvers or in response to complex weather conditions. To address these limitations, we proposed a PINN-based approach, leveraging first- and second-order numerical approximations implemented via forward finite differences. Our experimental results demonstrate that this approach provides a valuable trade-off, effectively combining data-driven learning with physical consistency, while exhibiting improved performance even with limited data.

While training PINNs can be computationally demanding and requires careful hyperparameter tuning, this approach presents a compelling direction towards more robust, accurate, and generalizable prediction models. To further advance this research, future work will integrate higher-order numerical approximations to improve prediction accuracy in complex maritime scenarios and over longer prediction horizons. The security and safety of maritime navigation are also inherently dependent on inter-vessel interactions. Consequently, our ongoing goal is to incorporate spatial relationships among vessels directly into our PINN framework, further ensuring physical plausibility in complex multi-vessel scenarios.

\begin{acks}
This research was partially supported by the \textit{National Council for Scientific and Technological Development} (CNPq 444325/2024-7), the Center for Artificial Intelligence (FAPESP 19/07665-4), and \textit{Dalhousie University} (DAL). The data used in this study was provided by AISViz/MERIDIAN and is subject to licensing restrictions, preventing the sharing of raw data. However, the pre-trained models and further code can be shared and are available upon request. 
\end{acks}
\balance
\bibliographystyle{unsrtnat}
\bibliography{references}
\end{document}